\documentclass[letterpaper]{article} 
\usepackage{aaai2026}  
\usepackage{times}  
\usepackage{helvet}  
\usepackage{courier}  
\usepackage[hyphens]{url}  
\usepackage{graphicx} 
\usepackage{natbib}  
\usepackage{caption} 
\usepackage{algorithm}
\usepackage{algorithmic}
\usepackage{multirow}
\usepackage{amsmath}
\usepackage{amssymb}
\usepackage{booktabs}
\usepackage{tabularx}

\usepackage{newfloat}
\usepackage{listings}
\DeclareCaptionStyle{ruled}{labelfont=normalfont,labelsep=colon,strut=off} 
\floatstyle{ruled}
\newfloat{listing}{tb}{lst}{}
\floatname{listing}{Listing}

\title{MCI-Net: A Robust Multi-Domain Context Integration Network for\\ Point Cloud Registration}
\author{
Shuyuan Lin\textsuperscript{\rm 1}, 
Wenwu Peng\textsuperscript{\rm 1}, 
Junjie Huang\textsuperscript{\rm 1}, 
Qiang Qi\textsuperscript{\rm 2}, 
Miaohui Wang\textsuperscript{\rm 3}, 
Jian Weng\textsuperscript{\rm 1}\thanks{Corresponding Author.} 
}

\affiliations{
\textsuperscript{\rm 1}College of Cyber Security, Jinan University, Guangzhou, China\\
\textsuperscript{\rm 2}School of Data Science, Qingdao University of Science and Technology, Qingdao, China\\
\textsuperscript{\rm 3}College of Computer Science \& Software Engineering, Shenzhen University, Shenzhen, China\\
swin.shuyuan.lin@gmail.com
}

\usepackage{bibentry}

\begin{document}

\maketitle

\begin{abstract}
	Robust and discriminative feature learning is critical for high-quality point cloud registration. However, existing deep learning-based methods typically rely on Euclidean neighborhood-based strategies for feature extraction, which struggle to effectively capture the implicit semantics and structural consistency in point clouds. To address these issues, we propose a multi-domain context integration network (MCI-Net) that improves feature representation and registration performance by aggregating contextual cues from diverse domains. Specifically, we propose a graph neighborhood aggregation module, which constructs a global graph to capture the overall structural relationships within point clouds. We then propose a progressive context interaction module to enhance feature discriminability by performing intra-domain feature decoupling and inter-domain context interaction. Finally, we design a dynamic inlier selection method that optimizes inlier weights using residual information from multiple iterations of pose estimation, thereby improving the accuracy and robustness of registration. Extensive experiments on indoor RGB-D and outdoor LiDAR datasets show that the proposed MCI-Net significantly outperforms existing state-of-the-art methods, achieving the highest registration recall of 96.4\% on 3DMatch. 
\end{abstract}

\begin{links}
	\link{Code}{http://www.linshuyuan.com}
\end{links}

\section{Introduction}
Point cloud registration seeks to estimate the optimal transformation between a pair of partially overlapping point clouds and serves as a fundamental task in various 3D vision applications~\cite{hu20242,zhang2024robust,lin2022coclustering}, such as reconstruction, tracking, and robotic localization. Classical methods typically adopt a two-stage paradigm: first, encoding geometric structures into feature descriptors; then, establishing correspondences by matching the most similar descriptors between the two frames.

Establishing robust and distinctive feature correspondences is essential for high-quality point cloud registration. Recently, deep neural network–based methods~\cite{ao2021spinnet,yu2023rotation} have been widely used for local feature extraction. However, most of these approaches rely on Euclidean geometry to define local neighborhoods—typically using k-nearest neighbors or fixed-radius search—which inherently limits their ability to capture high-level semantic and structural consistency.
\begin{figure}[t]
	\centering
	\includegraphics[width=1\linewidth]{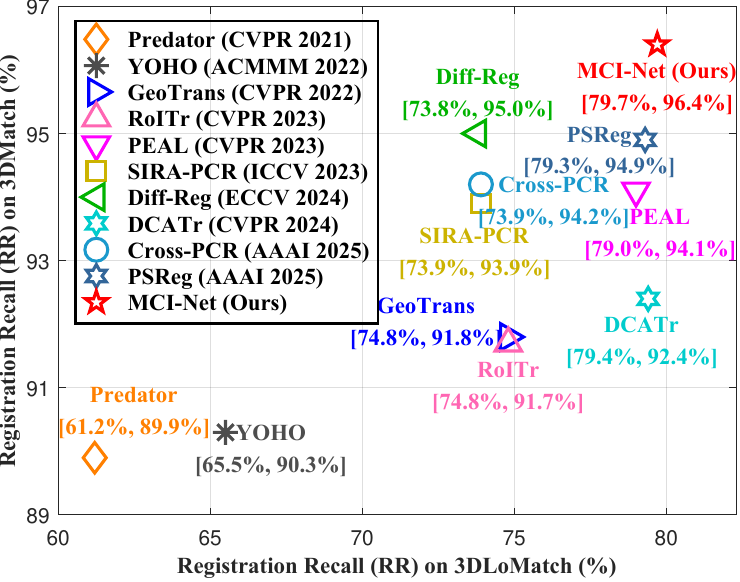}
	\caption{Registration recall (RR) on 3DLoMatch and 3DMatch with 2,500 points. The proposed method significantly outperforms all the state-of-the-art approaches.}
	\label{fig:figure1_3d_3dlo}
\end{figure}
\begin{figure*}[t]
	\centering
	\includegraphics[width=0.95\textwidth]{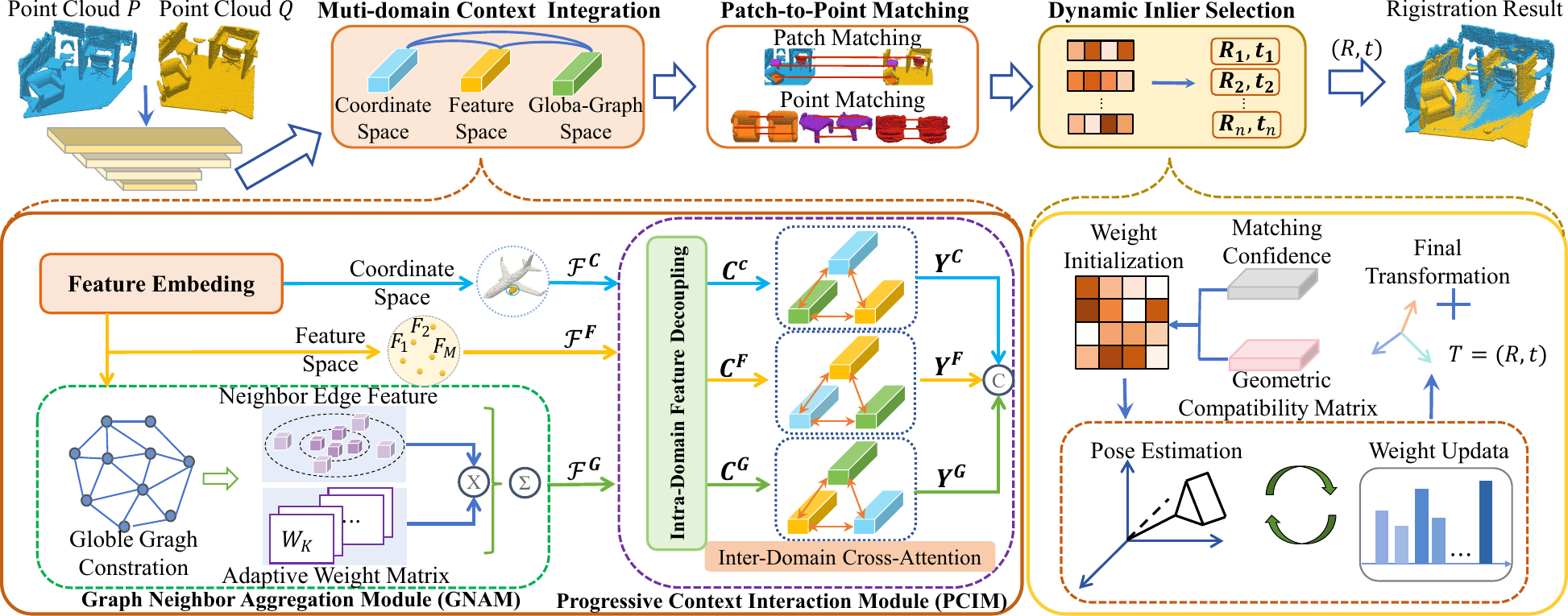}
	\caption{Architecture of the proposed MCI-Net. Given point clouds $P$ and $Q$, patch-level and point-level features are first extracted using the PAConv-based feature pyramid network~\cite{yao2024pare}. 
		The extracted patch features are embedded into coordinate and feature spaces, and then processed by the multi-domain context integration module, where GNAM captures global structural relationships, and PCIM decouples and fuses contextual information across domains to enhance feature discriminability.
		The patch-to-point matching strategy is used to obtain point correspondences. Finally, the dynamic inlier selection method iteratively updates correspondence weights to progressively refine the inlier set, yielding the final pose estimation $(R, t)$.}
	\label{fig:Overview-figure}
\end{figure*}
Moreover, relying solely on geometric neighborhoods to construct local structures in point clouds presents further limitations. These limitations hinder accurate feature extraction and the establishment of reliable correspondences, ultimately reducing the robustness and accuracy of registration.
To address these challenges, recent studies have employed hyperbolic feature embedding~\cite{xie2024hecpg,ermolov2022hyperbolic} to enhance the modeling of implicit semantic relationships in point clouds. Other works have introduced local attention mechanisms~\cite{lai2022stratified,fan2022svt,lai2022stratified,lin2024multistage} to assign varying levels of importance to each point's neighbors, thereby suppressing noise and emphasizing key features. Neighborhood learning methods, such as~\cite{lu2021hregnet}, use networks to learn the weights of neighbors within geometric clusters and aggregate them accordingly to improve representation quality. 
While these methods have improved the robustness and adaptability of local features, they remain constrained by geometric space modeling and fail to overcome the semantic expressiveness limitations of Euclidean neighborhoods.

Consequently, there remains a critical need for neighborhood construction strategies that can capture global structural patterns and implicit semantic relationships beyond local geometric constraints.
Inspired by recent advances in image matching~\cite{dai2022ms2dg,zhao2021progressive,liu2024ncmnet,luanyuan2024mgnet}, we propose the \textbf{multi-domain context integration network (MCI-Net)} for robust point cloud registration. 
First, we propose the \textbf{graph neighborhood aggregation module (GNAM)}  that effectively captures structural relationships within point clouds by constructing a global graph.
Second, we design the \textbf{progressive context interaction module (PCIM)} that combines intra-domain feature decoupling and inter-domain context interaction to enhance feature representation across spatial domains.
Finally, we propose the \textbf{dynamic inlier selection method (DISM)} that continuously optimizes the correspondence weights by integrating historical residuals from pose estimation, thereby adaptively highlighting reliable correspondences and suppressing outliers.
As shown in Figure~\ref{fig:figure1_3d_3dlo}, MCI-Net achieves the best RR compared to other competing methods in different overlapping scenes. 
An overview of the MCI-Net architecture is shown in Figure~\ref{fig:Overview-figure}, and the main contributions of this paper are as follows:
\begin{itemize}
	\item We propose MCI-Net, integrating multi-domain contextual information to effectively improve feature representation and registration performance.
	
	\item We propose GNAM that constructs a global graph to capture holistic structural relationships within point clouds.
	
	\item We propose PCIM that leverages intra-domain feature decoupling and intra-domain context interaction to enhance the representational and discriminative power of features through contextual information.
	
	\item We propose DISM that dynamically refines matching weights by incorporating historical residual information from multiple rounds of pose estimation, significantly improving registration accuracy and robustness.
\end{itemize}

\section{Related Work}
\subsection{3D Feature Descriptors}
Feature-based matching establishes stable point correspondences by comparing feature similarities between point clouds, without requiring prior pose information. In recent years, learning-based 3D descriptors have significantly outperformed traditional handcrafted methods. PPFNet~\cite{deng2018ppfnet} and PPF-FoldNet~\cite{deng2018ppf} utilize point pair features (PPF) to represent local structures and learn high-dimensional features through neural networks; yet these approaches still rely on handcrafted geometric relationships, limiting their adaptability to complex transformations such as rotations. To improve robustness under rotational variations, SpinNet~\cite{ao2021spinnet} maps local surfaces to cylindrical coordinates and extracts rotation-invariant local features using a 3D cylindrical convolutional network. Nevertheless, the expressive capability of cylindrical parameterization is limited, making it difficult to adapt to diverse geometric structures. YOHO~\cite{wang2022you} and RoReg~\cite{wang2023roreg} leverage group convolution based on the icosahedral group to construct rotation-equivariant and rotation-invariant feature descriptors. Due to the requirement of extracting features under 60 different rotations in the group structure, the overall computational cost is extremely high, and the method is less adaptable to non-uniformly distributed point clouds. Furthermore, RoITR~\cite{yu2023rotation} builds pose-invariant local geometric features and aggregates them via multiple attention layers to achieve rotation-invariant representations, but the use of multiple attention layers leads to high computational and memory overhead. PARE-Net~\cite{yao2024pare} employs position-aware rotation-equivariant convolution, explicitly incorporating spatial geometric information and rotation equivariance to enhance feature distinctiveness and robustness; however, it primarily focuses on modeling local neighborhood relationships, with limited capability for capturing global structural information across the entire point cloud.

Although existing methods have achieved notable progress in feature representation and robustness, most remain confined to Euclidean neighborhood modeling and fail to capture global structural relationships in complex scenarios. To overcome this limitation, this paper focuses on collaborative modeling across multiple spatial domains. By integrating geometric, feature, and global structural information, the proposed approach enhances structural perception and global consistency in point cloud matching, thereby significantly improving registration robustness.

\subsection{3D Transformation Estimators}
During the matching process between point pairs, the estimated correspondences often contain noise and outliers. As a result, a robust 3D transformation estimator is essential. The classical RANSAC algorithm~\cite{fischler1981random} is widely used due to its simplicity, but it suffers from limitations such as low accuracy and the need for many iterations to converge, which hinder its effectiveness in high-precision, real-time scenarios. DCP~\cite{wang2019deep} addresses this by using differentiable singular value decomposition (SVD) to estimate transformations based on soft correspondences derived from feature similarity. Its one-shot assignment strategy remains sensitive to ambiguous or noisy matches. More recently, coarse-to-fine correspondence learning methods have gained attention for their ability to refine matches in stages and improve robustness. CoFiNet~\cite{yu2021cofinet} introduces both self-attention and cross-attention layers in the coarse matching stage to learn descriptor correspondences, followed by optimal transport for fine matching. The matching process primarily relies on feature similarity and lacks geometric structural awareness. As a result, in scenarios with highly repetitive or symmetric structures, it is prone to significant matching ambiguity and a large number of outlier correspondences. GeoTrans~\cite{qin2022geometric} enhances local-to-global registration and pose estimation by embedding geometric structure into the self-attention mechanism to improve feature representation. However, when structural distinctiveness is weak or the data is noisy, the method struggles to filter out incorrect correspondences, which compromises registration robustness.

However, the aforementioned methods typically rely on correspondences generated in a single pass for pose estimation, making it difficult to effectively eliminate noise and outliers. In contrast, the strategy proposed in this paper adopts an iterative optimization approach. By incorporating a history-aware weight update mechanism, the method progressively refines the inlier set, leading to more robust and accurate 3D transformation estimation.

\section{Methodology}
Given two partially overlapping point clouds \( P = \{p_i \in \mathbb{R}^3 \mid i = 1, \dots, N\} \) and \( Q = \{q_j \in \mathbb{R}^3 \mid j = 1, \dots, M\} \), the point cloud registration problem seeks the optimal rigid transformation that minimizes the weighted sum of point-to-point errors of a predicted correspondence set \( C \), where each correspondence \( (p_k, q_k) \) is associated with a confidence weight \( w_k \). The transformation can be computed as follows:

\begin{equation}
	\min_{R, t} \sum_{(p_k, q_k) \in C} w_k \| R p_k + t - q_k \|^2_2,
\end{equation}
where \( \| \cdot \|_2 \) denotes the Euclidean norm, and \( R \in \mathrm{SO}(3) \) and \( t \in \mathbb{R}^3 \) represent the rotation and translation between \( P \) and \( Q \), respectively.

\subsection{Graph Neighborhood Aggregation}

To address the limitations of geometric neighborhood construction in capturing structural and semantic consistency in point clouds, we propose GNAM, which constructs a global graph using optimized graph convolutions and further performs adaptive neighborhood aggregation based on the constructed global graph.

\noindent\textbf{Global Graph Construction.} Specifically, Let \( \mathcal{F}^F \) denote the embedded features of each point generated by the feature embedding layer. These features are then fed into a prediction layer for probabilistic modeling, resulting in a local probability set \( \mathcal{W}_l \). We then define the global graph \( G_g = \{V_g, E_g\} \), where \( \mathcal{W}_l \) serves as the node set \( V_g \) of the global graph, and the edge set \( E_g \) consists of directed connections between each node and its spatial neighbors. Based on these node features, we construct a weighted adjacency matrix \( \tilde{A} \) via an outer product operation—i.e., using the correlation of \( \mathcal{W}_l \) to measure relationships between nodes in the graph—and incorporate an identity matrix \( \mathbf{I} \) to preserve self-connections. These operations can be expressed as:
\begin{equation}
	\mathcal{W}_l = \mathrm{ReLU}(\tanh(\text{MLP}(\mathcal{F}^F))),
\end{equation}
\begin{equation}
	\tilde{A} = \mathcal{W}_l \mathcal{W}_l^{\top} + \mathbf{I},
\end{equation}
where MLP(·) is a linear layer used to reduce the channel dimension to 1. The activation functions tanh(·) and ReLU(·) are used to compute the weights; a higher weight indicates a stronger correlation between connected points. Finally, we adopt a spectral graph convolutional layer~\cite{zhao2021progressive,kipf2016semi} as the global graph relationship modeler, yielding the global graph:
\begin{equation}
	F_g = \text{\(\sigma\)}(L\mathcal{F}^F),
\end{equation}
where \( L = \tilde{D}^{-\frac{1}{2}} \tilde{A} \tilde{D}^{-\frac{1}{2}} \) is the graph Laplacian used to project the original features into the spectral domain. \( \tilde{D} = \text{diag}\left(\sum_{j} \tilde{A}_{ij}\right) \) denotes the degree matrix of \( \tilde{A} \), and \( \sigma \) represents the ReLU(·) activation function.

\noindent\textbf{Adaptive Neighborhood Aggregation.} After constructing the global graph, it is necessary to effectively mine consensus information within neighborhood. Specifically, for each node \( p_i \), we construct a local neighborhood cluster \( \mathcal{G}_i = \{p_j\}_{j=1}^K \) in global graph based on the similarity principle of K-nearest neighbors. Standard pooling operators are typically used to aggregate features within the neighborhood cluster. However, such operations may discard important underlying relationships between graph nodes. To address this, we aggregate neighborhood information in the global graph using adaptive neighborhood weights:
\begin{equation}
	\quad \alpha_{ij} = \text{Softmax}(\text{MLP}(Concat[F_i, F_i - F_{ij}])),
\end{equation}
\begin{equation}
	\mathcal{F}^G_i = \sum_{p_j \in \mathcal{G}_i} \alpha_{ij} F_{ij},
\end{equation}
where \( {F}_i \) and \( {F}_{ij} \) are the feature representations of the central point \( p_i \) and its \( j \)-th spatial neighbor in the neighborhood cluster, respectively. The edge feature \(Concat[F_i, F_i - F_{ij}]\) of the neighborhood cluster is first passed through an MLP, followed by a Softmax function to compute the neighborhood attention weights \( \alpha_{ij} \). The neighborhood features \( F_{ij} \) are then weighted and summed using these attention weights to generate the computed aggregated feature \( \mathcal{F}^G_i \) for the central point \( i \). The final feature representation of the global graph is composed of the aggregated features \( \mathcal{F}^G_i \) for all points, denoted as \( \mathcal{F}^G \). This mechanism enables each node to effectively perceive features of surrounding points that are structurally consistent with it by adaptively aggregating neighborhood information in the global graph. As a result, it enhances both the structural consistency and discriminative power of the feature representation.

\subsection{Progressive Context Interaction}

To better exploit local-global contextual dependencies across multiple domains, we propose PCIM that enables contextual interaction at both intra-domain and inter-domain levels. At the intra-domain level, features are decomposed into components aligned with global features and residual components, allowing for the decoupling and joint modeling of local and global information. At the inter-domain level, we employ an inter-domain cross-attention to fuse contextual information from different domains, thereby further enhancing the representational capacity of features.

\noindent\textbf{Intra-Domain Feature Decoupling.}  
Let $\{\mathcal{F}^C, \mathcal{F}^F, \mathcal{F}^G\}$ denote the three feature sets from the coordinate space, feature space, and global graph, respectively, and let $\mathbf{C}_i$ represent a feature vector in a specific space domain. To construct global contextual information, we first perform global average pooling over all point features to obtain a global feature representation $\mathbf{C}_g$. To incorporate global context while preserving local structural information, we explicitly decompose each point's feature into a projection component aligned with the global feature and a residual component, thereby enabling structural decoupling and collaborative modeling of local and global information:
\begin{equation}
	\mathbf{C}_i^{\text{proj}} = \frac{\mathbf{C}_i \cdot \mathbf{C}_g}{\lVert \mathbf{C}_g \rVert^2} \mathbf{C}_g,
\end{equation}
\begin{equation}
	\mathbf{C}_i^{\text{res}} = \mathbf{C}_i - \mathbf{C}_i^{\text{proj}}.
\end{equation}
Here, $\mathbf{C}_i^{\text{proj}}$ denotes the projection component of the local feature onto the global feature, while $\mathbf{C}_i^{\text{res}}$ represents the residual component that captures local information independent of the global context. We then concatenate the residual component $\mathbf{C}_i^{\text{res}}$ with the global feature $\mathbf{C}_g$ to form the decoupled feature for the given space domain, constructing a unified representation that integrates both local structural integrity and global contextual awareness. The overall structure is illustrated in Figure \ref{fig:decoup}. Finally, we adopt a parallel architecture to independently model the features of each space domain, with the resulting features represented as $\{\mathbf{C}^C,\, \mathbf{C}^F,\, \mathbf{C}^G\} \in \mathbb{R}^{N \times d}$.

\noindent\textbf{Inter-Domain Context Interaction.}  
Since features from different domains capture diverse structural relationships and contextual information within the point cloud, we adopt a grouped approach to enable information complementarity and collaborative fusion across multi-domain features. As shown in Figure \ref{fig:Cross-attn}, we construct three parallel inter-domain cross-attention branches, each using one domain feature as the core while leveraging the other two to compute cross-attention. This enables selective information enhancement for the dominant feature. Taking the first branch as an example, its update process is formulated as:
\begin{equation}
	\Psi = \text{Softmax}\left( \phi_Q(\mathbf{C}^{F}) \cdot \phi_K(\mathbf{C}^{G})^\top \right),
\end{equation}
\begin{equation}
	\mathbf{Y}^C = \mathbf{C}^C + \lambda \cdot \mathrm{MLP}\left( \Psi \cdot \phi_V(\mathbf{C}^C) \right),
\end{equation}
where $\Psi$ denotes the attention weight matrix computed from the other two feature mappings, and $\phi_Q$, $\phi_K$, $\phi_V$ represent feature mapping modules consisting of linear layers, normalization, and activation functions. $\lambda$ is a learnable parameter initialized to 0. Similarly, the second and third branches produce the interaction features $\mathbf{Y}^F$ and $\mathbf{Y}^G$, respectively. Finally, we concatenate these three features along the channel dimension to form the overall output, achieving unified modeling of multi-neighborhood contextual information.
\begin{figure}[t]
	\centering
	\includegraphics[width=\linewidth]{figures/Fig3_Feture_Decoup2.pdf}
	\caption{The proposed intra-domain feature decoupling. Features are decomposed into global-aligned and residual parts, and the residual is then concatenated with the global feature to capture both local details and global context.}
	\label{fig:decoup}
\end{figure}
\subsection{Patch-to-Point Matching}
To construct the matching pipeline, we adopt a coarse-to-fine strategy~\cite{yu2021cofinet,qin2022geometric}, which divides the point cloud matching task into two stages. First, patch-level features are used for coarse filtering of the point clouds to identify potential overlapping regions, significantly reducing the search space for matching. Then, within the filtered candidate regions, fine-grained matching relationships are further estimated based on point-level features.

\noindent\textbf{Patch Matching.}  
For the source point cloud \(P\) and the target point cloud \(Q\), initial patch-level features \(\hat{\mathbf{X}}^{(P)}\) and \(\hat{\mathbf{X}}^{(Q)}\) are first extracted using the feature pyramid backbone. These features are then further enhanced by the multi-domain context integration module to obtain \(\hat{\mathbf{F}}^{(P)}\) and \(\hat{\mathbf{F}}^{(Q)}\), respectively. We use the GeoTrans method~\cite{qin2022geometric} to process these two sets of features, along with positional information, by passing them through alternating self-attention and cross-attention layers, thereby generating the mixed features \(\hat{\mathbf{H}}^{(P)}\) and \(\hat{\mathbf{H}}^{(Q)}\). A Gaussian correlation matrix $S_{m,n}$ is then computed for the mixed features. After normalization and top-\(k\) selection, we obtain the patch correspondence set $\hat{C} = \left\{ \left( \hat{\mathbf{p}}_{i}, \hat{\mathbf{q}}_{j} \right) \mid (i, j) \in \text{Topk}_{m,n} \left( S_{m,n} \right) \right\}$.

\noindent\textbf{Point Matching.}
After obtaining the patch correspondences $\hat{C} = (\hat{\mathbf{p}}_{i}, \hat{\mathbf{q}}_{j})$, we construct a similarity matrix based on the point-level features \(\mathbf{F}^{\hat{P}_i}\) and \(\mathbf{F}^{\hat{Q}_j}\) within the corresponding patch regions \((G_{\hat{P}_{i}}, G_{\hat{Q}_{j}})\):
\begin{equation}
	M_{i,j} = \frac{\mathbf{F}^{\hat{P}_i} \cdot \mathbf{F}^{\hat{Q}_j}}
	{\sqrt{\sum_{\hat{k}=1}^{\hat{d}} \left(\mathbf{F}^{\hat{P}_i}_{\hat{k}}\right)^2}
		\cdot
		\sqrt{\sum_{\hat{k}=1}^{\hat{d}} \left(\mathbf{F}^{\hat{Q}_j}_{\hat{k}}\right)^2}},
\end{equation}
where \(\hat{d}\) denotes the feature dimension. 
Then, the similarity matrix \(M_{i,j}\) is normalized using the Sinkhorn algorithm~\cite{sinkhorn1967concerning} to obtain a soft assignment matrix \(Z_i\). To ensure bidirectional reliability between point pairs, we retain only those point pairs that rank within the top-\(k\) scores in both the rows and columns of \(Z_i\), forming the point matching set \(\tilde{C_i}\) within the current patch. Finally, the local results of all \(N_c\) patch pairs are aggregated into the global point-level matching set \(\tilde{C} = \bigcup_{i=1}^{N_c} \tilde{C_i}\).

\noindent\textbf{Loss Function.}  
Following~\cite{yao2024pare}, we employs three loss functions to jointly optimize network training: the patch matching loss \( \mathcal{L}_{\text{patch}} \), the point matching loss \( \mathcal{L}_{\text{point}} \), and the contrastive rotation loss \( \mathcal{L}_{\text{rot}} \). The overall loss is defined as \( \mathcal{L} = \mathcal{L}_{\text{patch}} + \mathcal{L}_{\text{point}} + \mathcal{L}_{\text{rot}} \).

\begin{figure}[t]
	\centering
	\includegraphics[width=0.9\linewidth]{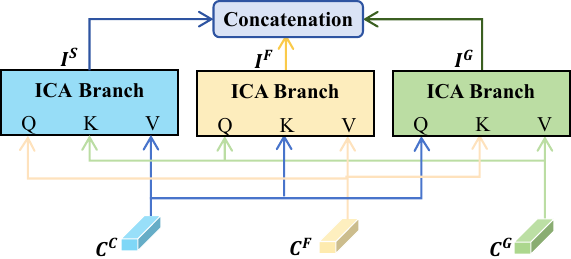}
	\caption{The proposed inter-domain context interaction. It contains three parallel inter-domain cross-attention (ICA) branches for contextual information fusion across domains.}
	\label{fig:Cross-attn}
\end{figure}

\subsection{Dynamic Inlier Selection}

Accurate pose estimation requires identifying reliable correspondences under ambiguous structures. To this end, we propose DISM to initialize weights jointly based on matching confidence and geometric consistency, and iteratively refine them using residual feedback.

\noindent\textbf{Weight Initialization.}  
Due to geometric similarities between local regions, multiple candidate matches with similar features often emerge, leading to matching ambiguities. To mitigate this, we introduce geometric compatibility as a simple yet effective consistency measure, based on the principle that the distance between point pairs should remain consistent before and after transformation~\cite{bai2021pointdsc,lee2021deep}. For each point-wise correspondence $(\tilde{\mathbf{p}}_{i}, \tilde{\mathbf{q}}_{j})$ established in the point matching set $\tilde{C}$, the initial weight \( w^{(0)}_{ij} \) is determined jointly by the matching confidence and the geometric compatibility matrix:
\begin{equation}
	w^{(0)}_{ij} = \beta_{ij} \cdot \gamma_{ij},\beta_{ij} = \max\left(0, 1 - \frac{d_{ij}^2}{\varepsilon^2} \right),
\end{equation}
where \( \gamma_{ij} \) denotes the confidence of correspondence. The geometric compatibility matrix \( \beta_{ij} \) measures the geometric consistency between matched correspondences. \( d_{ij} \) represents the distance difference between the source and target correspondence, and \( \varepsilon \) is a threshold parameter that controls the tolerance for length differences.

\noindent\textbf{History-Aware Iterative Pose Estimation.}  
To improve the robustness of pose estimation, we design an iterative optimization strategy incorporating a history-aware mechanism. In each iteration, we perform weighted SVD based on the current weighted consensus set to update the rigid transformation parameters. The weights of the correspondences are then adjusted according to the current rotation residuals, thereby reducing the influence of potential mismatches.
Specifically, let \( \hat{R}^{(n)} \) denote the rotation estimate obtained at the \( n \)-th iteration. The rotation residual between two consecutive estimates is expressed as:
\begin{equation}
	\theta^{(n)} = \text{Angle}\left( \hat{R}^{(n-1)}, \hat{R}^{(n)} \right),
\end{equation}
where Angle(·, ·) measures the angular difference between two rotation matrices, indicating the degree of change in the estimation. Since early iterations may produce unstable estimates, we introduce an iteration-dependent historical weighting mechanism to control the influence of each residual on the final weight update. The weight at the \( n \)-th iteration is updated as:
\begin{equation}
	w_{ij}^{(n)} = w_{ij}^{(0)} \cdot \prod_{k=1}^{n} \exp\left( - E(k) \cdot \theta^{(k)} \right),
\end{equation}
where  \( E(k) = 2k/(m(m+1)) \) is a normalized historical weight function that determines the influence of each residual, with \( m \) being the total number of iterations. The final transformation is represented by \( T = \{ R, t \} \), based on the pose estimate from the last iteration. This strategy incorporates multi-round residual information and mitigates the impact of unstable early estimates, thereby improving the accuracy and robustness of pose estimation.

\section{Experiments}
In this section, we evaluate the proposed method against various advanced methods on multiple benchmark datasets, including indoor RGB-D point cloud datasets 3DMatch~\cite{zeng20173dmatch} and 3DLoMatch~\cite{huang2021predator}, and outdoor LiDAR point cloud dataset KITTI Odometry~\cite{geiger2012we}. In addition, we have performed comprehensive ablation studies. All experiments are conducted on Ubuntu 18.04 with a single Nvidia RTX3090.

\begin{figure*}[t]
	\centering
	\includegraphics[width=\textwidth]{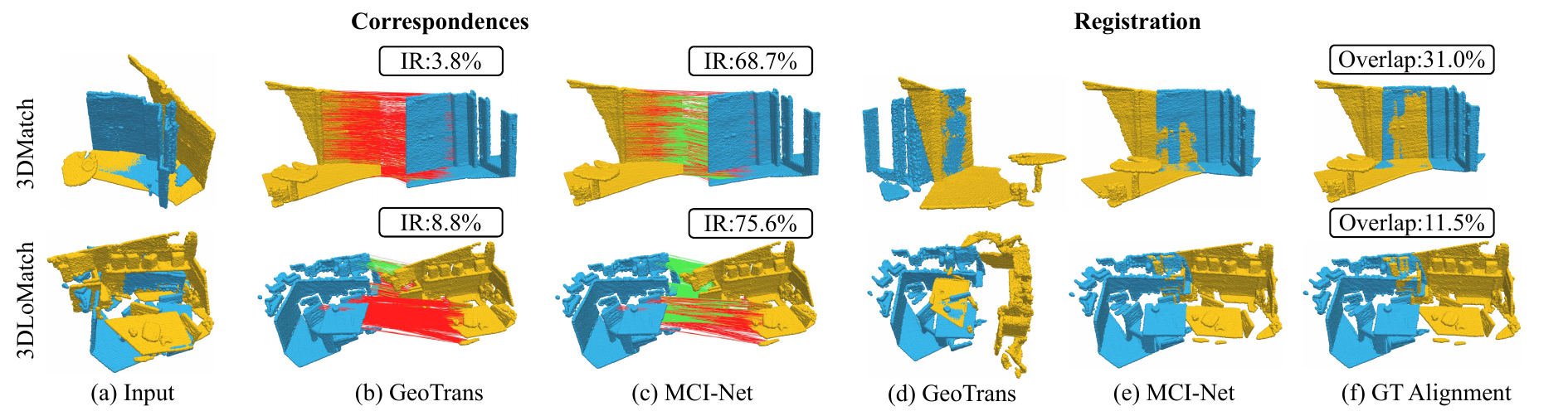}
	\caption{Qualitative results on 3DMatch and 3DLoMatch datasets. Columns (b) and (c) show the correspondences, while columns (d) and (e) demonstrate the registration results. Green/red lines indicate inliers/outliers.}
	\label{fig:visual}
\end{figure*}

\subsection{Indoor Scenarios: 3DMatch and 3DLoMatch}
\begin{table}[!t]
	\centering
	\fontsize{7pt}{8.2pt}\selectfont   
	\setlength{\tabcolsep}{3pt}      
	\renewcommand{\arraystretch}{0.93} 
	\begin{tabular}{l|ccccc|ccccc}
		\toprule
		& \multicolumn{5}{c|}{3DMatch} & \multicolumn{5}{c}{3DLoMatch} \\
		Samples & 5000 & 2500 & 1000 & 500 & 250 & 5000 & 2500 & 1000 & 500 & 250 \\
		
		\midrule
		\multicolumn{11}{c}{\textbf{Registration Recall (RR, \%)}} \\
		\midrule
		FCGF~ & 85.1 & 84.7 & 83.3 & 81.6 & 71.4 & 40.1 & 41.7 & 38.2 & 35.4 & 26.8 \\
		Predator~ & 89.0 & 89.9 & 90.6 & 88.5 & 86.6 & 59.8 & 61.2 & 62.4 & 60.8 & 58.1 \\
		YOHO~ & 90.8 & 90.3 & 89.1 & 88.6 & 84.5 & 65.2 & 65.5 & 63.2 & 56.5 & 48.0 \\
		GeoTrans~ & 92.0 & 91.8 & 91.8 & 91.4 & 91.2 & 75.0 & 74.8 & 74.2 & 74.1 & 73.5 \\
		RoITr~ & 91.9 & 91.7 & 91.8 & 91.4 & 91.0 & 74.7 & 74.8 & 74.8 & 74.2 & 73.6 \\
		PEAL~ & 94.4 & 94.1 & 94.1 & 93.9 & 93.4
		& 79.2 & 79.0 & \underline{78.8} & 78.5 & 77.9 \\
		SIRA-PCR~ & 93.6 & 93.9 & 93.9 & 92.7 & 92.4 & 73.5 & 73.9 & 73.0 & 73.4 & 71.1 \\
		Diff-Reg ~ &  &  & 95.0 &  &  &  &  & 73.8 &  &  \\
		DCATr ~ & 92.6 & 92.4 & 92.2 & 91.9 & 91.6 & 76.8 & 76.4 & 75.7 & 75.1 & 73.7 \\
		Cross-PCR ~ & 94.5 & 94.2 & 94.2 & 94.3 & 94.0 & 73.7 & 73.9 & 74.1 & 74.2 & 74.1 \\
		PSReg ~ & \underline{95.7} & \underline{94.9} & \underline{95.1} & \underline{95.0} & \underline{95.2} 
		& \underline{79.3} & \underline{79.3} & 78.7 & \underline{78.7} & \underline{78.4} \\
		MCI-Net & \textbf{96.4} & \textbf{96.4} & \textbf{96.6} & \textbf{96.5} & \textbf{96.6} 
		& \textbf{79.7} & \textbf{79.7} & \textbf{79.9} & \textbf{79.6} & \textbf{79.5} \\  
		
		\midrule
		\multicolumn{11}{c}{\textbf{Inlier Ratio (IR, \%)}} \\
		\midrule
		FCGF~ & 56.8 & 54.1 & 48.7 & 42.5 & 34.1 & 21.4 & 20.0 & 17.2 & 14.8 & 11.6 \\
		Predator~ & 58.0 & 58.4 & 57.1 & 54.1 & 49.3 & 26.7 & 28.1 & 28.3 & 27.5 & 25.8 \\
		YOHO~ & 64.4 & 60.7 & 55.7 & 46.4 & 41.2 & 25.9 & 23.3 & 22.6 & 18.2 & 15.0 \\
		GeoTrans~ & 71.9 & 75.2 & 76.0 & 82.2 & 85.1 & 43.5 & 45.3 & 46.2 & 52.9 & 57.7 \\
		RoITr~ & 82.6 & 82.8 & 83.0 & 83.0 & 83.0 & 54.3 & 54.6 & 55.1 & 55.2 & 55.3 \\
		PEAL~ & 74.8 & 81.3 & 86.0 & 87.9 & 89.2
		& 49.1 & 54.1 & 60.5 & 63.6 & 65.0 \\
		SIRA-PCR~ & 70.8 & 78.3 & 83.7 & 85.9 & 87.4 & 43.3 & 49.0 & 55.9 & 58.8 & 60.7 \\
		Diff-Reg ~ &  &  & 30.9 &  &  &  &  & 9.6 &  &  \\
		DCATr ~ & 87.6 & 86.5 & 84.7 & 81.0 & 76.5 
		& 62.1 & 60.3 & 57.9 & 53.3 & 48.4 \\
		Cross-PCR ~ & \underline{88.7} & \underline{88.7} & \underline{88.7} & 88.7 & 88.7
		& \underline{65.9} & \underline{65.9} & \underline{65.9} & \underline{65.9} & 65.9 \\
		PSReg ~ & 75.8 & 82.4 & 87.1 & \underline{88.9} & \underline{90.0} 
		& 49.9 & 55.5 & 61.9 & 64.5 & \underline{66.3} \\
		MCI-Net & \textbf{88.8} & \textbf{88.9} & \textbf{89.8} & \textbf{90.6} & \textbf{91.1} & \textbf{66.0} & \textbf{66.1} & \textbf{67.2} & \textbf{68.0} & \textbf{68.7} \\
		
		\midrule
		\multicolumn{11}{c}{\textbf{Feature Matching Recall (FMR, \%)}} \\
		\midrule
		FCGF~ & 97.4 & 97.3 & 97.0 & 96.7 & 96.6 & 76.6 & 75.4 & 74.2 & 71.7 & 67.3 \\
		Predator~ & 96.6 & 96.6 & 96.5 & 96.3 & 96.5 & 78.6 & 77.4 & 76.3 & 75.7 & 75.3 \\
		YOHO~ & 98.2 & 97.6 & 97.5 & 97.7 & 96.0 & 79.4 & 78.1 & 76.3 & 73.8 & 69.1 \\
		GeoTrans~ & 97.9 & 97.9 & 97.9 & 97.9 & 97.6 
		& 88.3 & 88.6 & 88.8 & \underline{88.6} & \underline{88.3} \\
		RoITr~ & 98.0 & 98.0 & 97.9 & 98.0 & 97.9 & \textbf{89.6} & \textbf{89.6} & \textbf{89.5} & \textbf{89.4} & \textbf{89.3} \\
		PEAL~ & \underline{98.4} & \underline{98.4} & \underline{98.4} & 98.4 & 98.4
		& 87.7 & 87.8 & 87.8 & 88.0 & 87.4 \\
		SIRA-PCR~ & 98.2 & \underline{98.4} & \underline{98.4} & \underline{98.5} & \underline{98.5} 
		& \underline{88.8} & \underline{89.0} & \underline{88.9} & \underline{88.6} & 87.7 \\
		Diff-Reg ~ &  &  & 96.2 &  &  &  &  & 69.6 &  &  \\
		DCATr ~ & 98.2 & 98.3 & \underline{98.4} & 98.0 & 98.1 & 87.7 & 87.5 & 87.7 & 87.2 & 87.4 \\
		Cross-PCR ~ & 97.9 & 97.7 & 97.9 & 97.7 & 97.7 & 83.5 & 83.3 & 83.1 & 83.1 & 83.4 \\
		PSReg ~ & \textbf{98.6} & \textbf{98.6} & \textbf{98.6} & \textbf{98.6} & \textbf{98.6}
		& 86.4 & 86.6 & 87.1 & 87.4 & 87.1 \\
		MCI-Net & \underline{98.4} & \underline{98.4} & \underline{98.4} & 98.4 & \underline{98.5} & 85.1 & 85.1 & 85.1 & 85.1 & 85.0 \\
		\bottomrule
	\end{tabular}
	\caption{Evaluation results on 3DMatch and 3DLoMatch. The best and second-to-best results of baseline methods are respectively marked in bold and underlined.}
	\label{tab:3dmatch}
\end{table}


\noindent\textbf{Dataset.}  
The proposed MCI-Net is evaluated on the indoor point cloud registration benchmark datasets 3DMatch~\cite{zeng20173dmatch} and 3DLoMatch. According to~\cite{huang2021predator}, 3DMatch consists of point cloud pairs with overlap ratios above 30\%, while 3DLoMatch contains pairs with overlap ratios ranging from 10\% to 30\%.

\noindent\textbf{Metrics.}  
Following~\cite{bai2020d3feat,qin2022geometric}, we report inlier ratio (IR), feature matching recall (FMR), and registration recall (RR) as evaluation metrics.

\noindent\textbf{Registration Results.}  
%
In Table \ref{tab:3dmatch}, we compare the correspondence results of MCI-Net with those of other recent methods, such as FCGF~\cite{choy2019fully}, Predator~\cite{huang2021predator}, YOHO~\cite{wang2022you}, GeoTrans~\cite{qin2022geometric}, RoITr~\cite{yu2023rotation}, PEAL~\cite{yu2023peal}, SIRA-PCR~\cite{chen2023sira}, Diff-Reg~\cite{wu2024diff}, DCATr~\cite{chen2024dynamic}, Cross-PCR~\cite{zhao2025cross}, and PSReg~\cite{huang2025psreg}, under varying numbers of correspondences (i.e., 5000, 2500, 1000, 500 and 250). 
Compared to these methods, MCI-Net demonstrates significant improvements in RR and IR on both the 3DMatch and 3DLoMatch datasets. In terms of RR, this metric directly reflects the registration success rate and is often considered the most critical metric. MCI-Net consistently achieves the highest RR under varying sampling densities, owing to its ability to learn contextual information from multiple domain spaces, effectively addressing the limitations of methods that rely solely on geometric positions for feature extraction. In terms of IR, MCI-Net outperforms all existing methods across all sampling density settings. This improvement is largely attributed to our DISM, which dynamically increases the weights of high-confidence inliers during the iterative process, preserving true correspondences while suppressing noise and mismatches. In terms of FMR, MCI-Net performs comparable to PSReg and PEAL on the 3DMatch dataset. It is important to note that PSReg and PEAL require additional prior information about overlap, whereas MCI-Net operates without any prior information. However, in extremely low-overlap scenarios (e.g., 3DLoMatch), the sparse nature of the data makes it challenging to extract sufficient structural cues and reliable inliers, which results in a decrease in the FMR metric for MCI-Net.

\noindent\textbf{Qualitative Results.}
We present qualitative results in Figure~\ref{fig:visual}, where our results are compared with GeoTrans~\cite{qin2022geometric}. The results show that MCI-Net achieves higher inlier ratios (IR) for correspondences on both the 3DMatch and low-overlap 3DLoMatch datasets, and demonstrates more accurate registration results.


\begin{table}[!t]
	\centering
	\fontsize{7pt}{8.2pt}\selectfont   
	\setlength{\tabcolsep}{6pt}      
	\renewcommand{\arraystretch}{0.93} 
	\begin{tabular}{l|ccc|c}
		\toprule
		Methods & RRE ($^\circ$) & RTE (cm) & RR (\%) & Time (s)\\
		\midrule
		FCGF     & 0.30 & 9.5 & 96.6 & \textbf{0.18}\\
		D3Feat   & 0.30 & 7.2 & \textbf{99.8} & -\\
		Predator & 0.27 & 6.8 & \textbf{99.8} & 0.69\\
		SpinNet  & 0.47 & 9.9 & \underline{99.1} & 14.57\\
		CoFiNet  & 0.41 & 8.2 & \textbf{99.8} & 0.65\\
		BUFFER   & \underline{0.26} & 7.1 & \textbf{99.8} & 0.30\\
		GeoTrans & \textbf{0.23} & \underline{6.2} & \textbf{99.8} & 0.31\\
		MCI-Net  & \textbf{0.23} & \textbf{5.0} & \textbf{99.8} & \underline{0.23}\\
		\bottomrule
	\end{tabular}
\caption{Evaluation results on KITTI Odometry. The best and second-to-best results of baseline methods are respectively marked in bold and underlined.}
\label{tab:kitti}
\end{table}

\subsection{Outdoor Scenarios: KITTI Odometry}

\noindent\textbf{Dataset.}  
KITTI Odometry~\cite{geiger2012we} is a widely used outdoor benchmark dataset captured by LiDAR sensors for autonomous driving applications. Following~\cite{bai2020d3feat,qin2022geometric}, we use sequences 0–5 for training, 6–7 for validation, and 8–10 for testing.

\noindent\textbf{Metrics.}  
Following prior work~\cite{huang2021predator}, we employ three metrics to evaluate: relative rotation error (RRE), relative translation error (RTE) and registration recall (RR).

\noindent\textbf{Registration Results.}  
We evaluate our performance by comparing experimental results with recent methods, including FCGF~\cite{choy2019fully}, D3Feat~\cite{bai2020d3feat}, Predator~\cite{huang2021predator}, SpinNet~\cite{ao2021spinnet}, CoFiNet~\cite{yu2021cofinet}, BUFFER~\cite{ao2023buffer}, and GeoTrans~\cite{qin2022geometric}. The quantitative results are shown in Table~\ref{tab:kitti} and demonstrate that MCI-Net achieves leading performance on the KITTI Odometry dataset. Specifically, MCI-Net achieves the highest RR, as well as the lowest RRE and RTE among all compared methods. Furthermore, MCI-Net delivers outstanding runtime performance. Although FCGF achieves slightly faster runtime than MCI-Net, its RR is 3.2\% lower than MCI-Net. The experimental results demonstrate that MCI-Net generalizes well to outdoor datasets.

\begin{table}[!t]
\centering
\fontsize{7pt}{8.2pt}\selectfont   
\setlength{\tabcolsep}{5pt}      
\renewcommand{\arraystretch}{0.93} 
\begin{tabular}{c|l|ccc|ccc}
	\toprule
	&  & \multicolumn{3}{c|}{3DMatch} & \multicolumn{3}{c}{3DLoMatch} \\
	No. & Methods & RR & IR & FMR & RR & IR & FMR \\
	\midrule
	(1) & Full Model (Ours) & \textbf{96.4} & \textbf{88.8} & \textbf{98.4} & \textbf{79.7} & \textbf{65.9} & \textbf{85.1} \\
	(2) & w/o GNAM  & 95.2 & 87.9 & 98.0 & 78.9 & 65.8 & 85.0 \\
	(3) & w/o IFD  & 95.3 & 88.1 & 98.1 & 79.3 & 65.8 & \textbf{85.1} \\
	(4) & w/o ICI  & 95.7 & 88.3 & 98.3 & 79.5 & 65.7 & 85.0 \\
	\bottomrule
\end{tabular}
\caption{Ablation studies of MCI-Net.}
\label{tab:ablation_study}
\end{table}

\subsection{Ablation Studies}  
\noindent\textbf{Graph Neighborhood Aggregation Module (GNAM):} 
As shown in Table~\ref{tab:ablation_study} (2), we remove GNAM results in a noticeable drop in performance. This validates the effectiveness of our approach in enhancing global structural awareness and improving registration robustness.
\\\noindent\textbf{Intra-Domain Feature Decoupling (IFD):}  
As shown in Table~\ref{tab:ablation_study} (3), we remove the intra-domain feature decoupling module, which limits the ability of the model to fully exploit the synergy between local structure and global context. This results in limited feature representation and a substantial decline in registration accuracy. 
\\\noindent\textbf{Inter-Domain Context Interaction (ICI):}  
We replace the intra-domain cross-attention with independent self-attention in each domain, thereby removing explicit fusion among features from different domains. As shown by the performance drop in Table~\ref{tab:ablation_study} (4), this ablation demonstrates that employing intra-domain cross-attention to fuse contextual information from different domains further enhances feature representation and registration robustness.
\\\noindent\textbf{Pose Estimator:}  
To demonstrate the superiority of our proposed pose estimator, we compare our DISM with two classical baselines: RANSAC~\cite{fischler1981random} and LGR~\cite{qin2022geometric}. As shown in Table~\ref{tab:pose_ablation}, MCI-Net achieves consistently competitive performance across all evaluation metrics on both the 3DMatch and 3DLoMatch datasets. These results validate that, compared to traditional pose estimation methods, the proposed method offers significantly better robustness and accuracy.

\begin{table}[!t]
\centering
\fontsize{7pt}{8.2pt}\selectfont   
\setlength{\tabcolsep}{5pt}      
\renewcommand{\arraystretch}{1} 
\begin{tabularx}{\linewidth}{l|*{3}{>{\centering\arraybackslash}X}|*{3}{>{\centering\arraybackslash}X}}
\hline
& \multicolumn{3}{c|}{3DMatch} & \multicolumn{3}{c}{3DLoMatch} \\
Pose Estimator & RR & RRE & RTE & RR & RRE & RTE \\
\hline
RANSAC & 94.6 & 2.29 & 0.072 & 78.8 & 3.51 & 0.103 \\
LGR & 94.9 & 1.88 & 0.060 & 79.1 & 2.88 & 0.086 \\
DISM (Ours) & \textbf{96.4} & \textbf{1.72} & \textbf{0.054} & \textbf{79.8} & \textbf{2.63} & \textbf{0.075} \\
\hline
\end{tabularx}
\caption{Ablation studies of pose estimator.}
\label{tab:pose_ablation}
\end{table}

\section{Conclusion}
In this paper, we propose MCI-Net, the multi-domain context integration network designed to enhance feature representation and registration performance. To capture the overall structural relationships of point clouds, we propose GNAM within the constructed global graph. To enhance the contextual representation and discriminative capability of features, we propose PCIM, which decouples intra-domain features and fuses inter-domain contexts across spatial scales. Moreover, we propose DISM that iteratively refines inlier weights by incorporating historical residuals from pose estimation, effectively enhancing registration reliability and robustness. Extensive experiments on indoor and outdoor benchmarks show that MCI-Net significantly outperforms recent competing methods. However, in extremely low-overlap scenarios, MCI-Net still faces certain limitations in registration performance due to the scarcity of structural cues and reliable inliers. In future work, we plan to further explore its potential and scalability in large-scale point cloud scenarios and more diverse processing tasks.

\clearpage
\section{Acknowledgments}
This work was supported in part by National Natural Science Foundation of China (Nos. U22A2095, 62476112, 62332007, U22B2028, 62501343); in part by Guangdong Basic and Applied Basic Research Foundation (Nos. 2024A1515011740, 2025A1515010181); 
in part by Natural Science Foundation of Shandong Province (No. ZR2024QF294); 
in part by Fundamental Research Funds for the Central Universities (Nos. 21624404, 23JNSYS01); 
in part by Science and Technology Major Project of Tibetan Autonomous Region of China (No. XZ202201ZD0006G), National Joint Engineering Research Center of Network Security Detection and Protection Technology, Guangdong Key Laboratory of Data Security and Privacy Preserving, Guangdong Hong Kong Joint Laboratory for Data Security and Privacy Protection, Engineering Research Center of Trustworthy AI, Ministry of Education, and Guangdong Key Laboratory of Intelligent Information Processing, Shenzhen University.

\bibliography{aaai2026}

\end{document}